\documentclass{article} 
\usepackage{iclr2026_conference,times}

\usepackage{amsmath,amsfonts,bm}

\def\eqref#1{equation~\ref{#1}}

\def\1{\bm{1}}

\DeclareMathAlphabet{\mathsfit}{\encodingdefault}{\sfdefault}{m}{sl}
\SetMathAlphabet{\mathsfit}{bold}{\encodingdefault}{\sfdefault}{bx}{n}

\newcommand{\R}{\mathbb{R}}

\usepackage{hyperref}
\usepackage{url}
\usepackage{graphicx}
\usepackage{wrapfig}
\usepackage{booktabs} 
\usepackage{siunitx}

\newcommand\bs[1]{\boldsymbol{#1}}

\title{Double projection for reconstructing dynamical systems: between stochastic and deterministic regimes}

\author{
  Viktor Sip, Martin Breyton, Spase Petkoski \& Viktor Jirsa  \\
  Aix Marseille Univ, INSERM, INS, Inst Neurosci Syst, Marseille, France \\
  \texttt{\{viktor.sip,martin.breyton,spase.petkoski,viktor.jirsa\}@univ-amu.fr} \\
}

\iclrfinalcopy 
\begin{document}

\maketitle

\begin{abstract}
Learning stochastic models of dynamical systems from observed data is of interest in many scientific fields.
Here, we propose a new method for this task within the family of dynamical variational autoencoders.
The proposed double projection method estimates both the system state trajectories and the noise time series from data.
This approach naturally allows us to perform multi-step system evolution and to learn models with a comparatively low-dimensional state space.
We evaluate the performance of the method on six benchmark problems, including both simulated and experimental data.
We further illustrate the effects of the teacher forcing interval of the multi-step scheme on the nature of the internal dynamics and compare the resulting behavior to that of deterministic models of equivalent architecture.
\end{abstract}

\section{Introduction}
\label{introduction}

Many scientific fields are concerned with building mathematical models of dynamical systems underlying the observed data.
Over the last decade, works using artificial neural networks to achieve this goal in a data-driven fashion have emerged, showing considerable promise \citep{DurstewitzEtAl23,LegaardEtAl23}.
Compared with the related task of time series prediction, the problem of dynamical system reconstruction (DSR) is differentiated by the following aspects: focus on the long-term dynamics of the trained system, an interest in the interpretable structure of the variables and the latent space, and reasoning and analysis of the learned system by the conceptual and computational tools from dynamical system theory.

Many of the influential DSR approaches assume that the underlying dynamics is deterministic \citep{BruntonEtAl16,PandarinathEtAl18,ChenEtAl18a,HessEtAl23}.
Indeed, finding a deterministic model that can accurately predict future behavior, matches the long-term properties observed in the data, and correctly generalizes to new conditions can be viewed as the ultimate goal of DSR.
To capture the long-term dependencies and patterns present in the data, the training methods for deterministic models often use multi-step prediction.
Such an approach, however, can lead to exploding or vanishing gradients and unstable training, especially for chaotic systems. To alleviate the problem, teacher forcing strategies for dynamical system reconstruction have been developed and analyzed \citep{MikhaeilEtAl22,HessEtAl23}.

Finding the deterministic system underlying the data is often infeasible due to the complexity of the generating process or limited experimental data.
Furthermore, reconstructing a deterministic model might not even be desirable if the resulting model is too computationally demanding for the intended purposes.
In such cases, relying on a stochastic framework might be a preferred alternative.
Stochastic models include a source of noise in the dynamical equations, in addition to the noisy observations that are commonly used with deterministic models.
An influential family of methods for training models of stochastic systems are Dynamical Variational Autoencoders \citep[DVAE,][]{GirinEtAl21}, a specific form of variational autoencoders \citep[VAE,][]{KingmaWelling14} for time series data.
A multi-step prediction with DVAE methods requires further adaptation of the framework.
As a notable example, a \textit{latent overshooting} approach \citep{HafnerEtAl19} was developed by modifying the cost function to use an $n$-step predictive distribution.
Using this method for multi-step prediction, the noise has to be sample from the prior distribution of the stochastic variables.

In this work, we propose a new variant of the DVAE approach using a double projection strategy, where we map the observations to both the system states and noise time series, on which we train the dynamical model.
It has several nice properties: 
First, it allows us to train models with comparatively low-dimensional state space, more suitable for analysis than high-dimensional recurrent neural networks used in other DVAE approaches.
Second, its special case is a fully deterministic method, allowing investigation of the role of stochasticity by comparison of the stochastic and deterministic variants.
Third, it allows us to perform multi-step prediction by sampling the noise from the posterior distribution, which we hypothesize can lead to better performance.

To evaluate the performance of the method, we test the method on six test problems, including models of both deterministic chaos and noise-driven dynamics, and experimental data.
Then, we investigate the effect of the teacher forcing interval on the nature of resulting dynamics, showing that the low values of the teacher forcing interval lead to deterministic dynamics, while large values result in a stochastic regime influenced by noise.

\section{Related work}

\paragraph{Teacher forcing.} The idea of teacher forcing for training deterministic recurrent neural networks (RNNs) consists of replacing the generated system states by the observed ones at predefined intervals in order to ensure that the generated trajectory stays close to the data.
The ideas emerged early in the context of training RNNs \citep{WilliamsZipser89,Doya92}.
Adaptive schemes with variable teacher forcing intervals were later developed \citep{BengioEtAl15}.
In the context of dynamical system reconstruction of chaotic systems, the approach was analyzed by \citet{MikhaeilEtAl22}.
A variant named generalized teacher forcing, wherein the states are interpolated between the generated and teacher system states, was developed by \citet{HessEtAl23}.

\paragraph{Stochastic models for DSR.} A range of methods to train a stochastic dynamical system from data was developed and explored in recent years.
A commonly used approach for this task is Dynamical Variational Autoencoders \citep{GirinEtAl21}, which merges the probabilistic state space models with the framework of variational autoencoders \citep{KingmaWelling14}.
The key components of the approach are an encoder network that maps the observations into the time series of latent variables, a discrete-time state space model parameterized by a flexible neural network, a decoder that maps from the latent space back to the observations, and a training method based on minimizing the evidence lower bound (ELBO).
The pioneering works often focused on other applications than DSR, among them are Deep Kalman Filters \citep{KrishnanEtAl15,KrishnanEtAl17} with latent variables being corresponding to the states of the system, Stochastic Recurrent Networks \citep{BayerOsendorfer15} with latent variables representing the noise time series, or Variational Recurrent Neural Networks \citep{ChungEtAl16} and Recurrent State Space Models \citep{HafnerEtAl19}, which combine deterministic and stochastic states.
In the explicit context of DSR,
\citet{KramerEtAl22} used the VAE framework to integrate multimodal data.
\citet{HernandezEtAl20} trained state-dependent linear networks, reusing the generative model inside the recognition model.
\citet{SipEtAl23} used coupled stochastic models to learn a model of brain network dynamics.

\paragraph{Multi-step prediction in stochastic models.} Several approaches have been developed to facilitate training with multi-step prediction for methods from the DVAE family. \citet{AmosEtAl18} introduced a multi-step deterministic rollout, comparing the predicted and actual observations (\textit{observation overshooting}).
Noting that projecting the system states to observations at every step can be costly for 2D image observations, \citet{HafnerEtAl19} compared the predictions in latent space (\textit{latent overshooting}) using multi-step predictions with noise sampled from the prior distribution.
\citet{LiEtAl21} investigated both observation and latent overshooting, and used prior distribution for noise samples.

\paragraph{Non-VAE stochastic models for DSR.}
Different methods than dynamical VAEs for training neural network based models of stochastic dynamics exist.
\citet{KoppeEtAl19} used the expectation-maximization (EM) method, where the latent states of the dynamical model are optimized directly, as opposed to through an encoder. 
\citet{KramerEtAl22} compared the EM method with a VAE-based approach, concluding better performance of EM compared to VAE for smaller problems, but at a cost of limited flexibility.
\citet{PalsEtAl24} trained stochastic low-rank recurrent neural networks using the variational sequential Monte Carlo method \citep{NaessethEtAl18}.
Other works use a parameterization other than neural networks for the stochastic generative model.
\citet{LindermanEtAl17} rely on a collection of linear dynamical systems with state-dependent probabilistic switching between them, leading to an interpretable representation.
Using Gaussian processes to represent the dynamics of the generative model \citep{DoerrEtAl18,GarciaEtAl22} naturally links the notion of uncertainty of the dynamics, but it requires careful choice of the kernel and is limited to low-dimensional state spaces.

\section{Methods}
\label{sec:methods}

\subsection{Double Projection Dynamical System Reconstruction}

In this study, we consider a dataset of $N$ system observations with dimension $d_x$ over $T$ time steps ${\{ \bs{x}_{1:T}^{(i)} \in \R^{T \times d_x} \}_{i=1}^N }$.
Our aim is to learn an underlying dynamical system 
\begin{align*}
  z_{t} & = F(z_{t-1}, \epsilon_t ), \\
  x_{t} & = g(z_{t}) + \Sigma_\eta \eta_t,
\end{align*}
with system states $z_t \in \R^{d_z}$, system noise $\epsilon_t \in \R^{d_\epsilon}$ of possibly lower dimension $d_\epsilon \leq d_z$, and observation noise $\eta_t \in \R^{d_x}$.

For this goal, we introduce a new variant of VAE-based methods.
Compared to existing VAE-based approaches, it combines two convenient features.
First, it separates computational expressivity from the size of the latent state, leading to comparatively low-dimensional latent spaces, more convenient for the analysis of its dynamical structures.
Second, it allows us to naturally perform multi-step prediction using samples from posterior distribution of the noise.
The existing approaches either use only a single-step prediction $p(z_t | z_{t-1})$ (e.g. Deep Kalman Filter \citep{KrishnanEtAl15,KrishnanEtAl17}), perform multi-step rollout using deterministic dynamics \citep{AmosEtAl18}, or sample the noise in multi-step rollout from the prior distribution of the noise \citep{HafnerEtAl19,LiEtAl21}.
We speculate that samples from the prior distributions are insufficient to approximate the high-dimensional integrals associated with multi-step rollout. Our proposed method, on the other hand, multi-step rollout can be performed with samples from posterior distribution of the noise.

\begin{figure}[t]
  \begin{center}
    \includegraphics[width=0.7\textwidth]{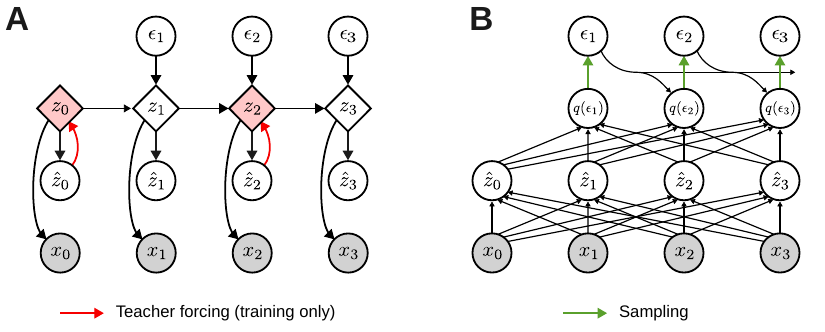}
  \end{center}
  \caption{Graphical summary of the DPDSR method.
    (A) Generation, visualized for teacher forcing interval $\tau = 2$.
    (B) Encoding.
    For brevity, we use a shorthand for the posterior distributions $q(\epsilon_t)$ instead of $q(\epsilon_t \mid \bs{x}, \hat{\bs{z}}, \bs{\epsilon}_{1:t-1})$.
  }
  \label{fig:method}  
\end{figure}

To achieve this, our method uses trained encoders to estimate both the system states and the system noise. We name the method Double Projection Dynamical System Reconstruction (DPDSR, Fig.~\ref{fig:method}).
Given one sample of the observed time series $\bs{x}_{1:T}$, it first estimates the (possibly partial) state space trajectory $\hat{\bs{z}}_{1:T}$, and subsequently also the noise time series $\bs{\epsilon}_{1:T}$.
Then, starting from the estimated initial conditions, the system is evolved according to the trained dynamical system and using the estimated noise time series. Every $\tau$ steps, the state of the dynamical system is set to the estimated state $\hat{z}_t$. To calculate the loss function, the match of the generated trajectories to the observations $\bs{x}_{1:T}$ and the estimated state space trajectory $\hat{\bs{z}}_{1:T}$ is combined with the Kullback-Leibler (KL) divergence of the latent variables $\bs{\epsilon}_{1:T}$ from the white noise prior.

In the following text, we write $\bs{x} = \bs{x}_{1:T}$ (and analogously for other variables) for readability.

\paragraph{Generative model}

We consider a generative model of the following form:
\begin{align}  
  z_{t} & = \tanh (\ f(z_{t-1}) + B \epsilon_t \ ) , \label{eq:gen-model} \\
  x_{t} & = g(z_{t}) + \Sigma_\eta \eta_t . \notag
\end{align}
The evolution function has the form of a two-layer perceptron with residual connection,
\begin{equation}
  f(z_t) = z_t + W_2 \sigma(W_1 z_t + b_1) + b_2, \label{eq:fz}
\end{equation}
with $\sigma(z)$ being the ReLU function.
The tanh nonlinearity is added to stabilize the dynamics and training by constraining the states to a finite interval.
The observation function is a two-layer perceptron, $g(z_t) = W^g_2 \sigma(W^g_1 z_t + b^g_1) + b^g_2$.
In the examples presented here, we use a one-dimensional noise ($d_\epsilon = 1$) injected into the last dimension only, so that $B = [0, \ldots, 0, \sigma_\epsilon^2]^T \in \R^{d_z \times 1}$.
The observation covariance matrix is diagonal and isotropic, $\Sigma_\eta = \sigma_\eta^2 I$.

\paragraph{Encoder}

The encoding process has two steps.  First, from the observed timeseries $\bs{x} \in \R^{T \times d_x}$ we compute a deterministic estimation of the system state timeseries $\hat{\bs{z}} \in \R^{T \times d_{\hat{z}}}$. These can be possibly only partial estimation of some dimensions of the states, ${d_{\hat{z}} \leq d_z}$.
We use a WaveNet architecture \citep{OordEtAl16}, which is based on 1D dilated convolutional networks. Unless specified otherwise, we use a single stack of seven dilated convolutional layers (Tab.~\ref{tab:params-enc}).
The output of the last layer is linearly projected to an estimation  $\hat{\bs{z}}$.
We are using non-causal layers, although we also train an auxilliary causal stack, which we use for prediction tasks (Sec.~\ref{sec:causal-encoder}).

In the second step, we estimate the posterior distribution $q(\bs{\epsilon} \mid \bs{x}, \hat{\bs{z}})$ from the observed time series $\bs{x}$ and estimated states $\hat{\bs{z}}$. The noise $\bs{\epsilon}$ serve as the latent variable in our VAE framework.
We use an autoregressive Gaussian posterior, $q(\epsilon_t \mid \bs{x}, \hat{\bs{z}}, \bs{\epsilon}_{1:t-1}) = N(\epsilon_t \mid \mu_t(\bs{x}, \hat{\bs{z}}, \bs{\epsilon}_{1:t-1}), \sigma_t(\bs{x}, \hat{\bs{z}}, \bs{\epsilon}_{1:t-1}))$. The means $\mu_t$ and variances $\sigma_t$ are computed by passing the input timeseries through a WaveNet block with the same architecture as in the first step, followed by an autoregressive LSTM with probabilistic output. This autoregressive form of the posterior distribution serves to increase the expressivity of the encoder of the posterior distribution.  

\paragraph{Training and loss function}

For each time series $\bs{x}$, the data is first projected to estimate of state $\hat{\bs{z}}$ and noise $\bs{\epsilon}$ via the described encoders.
If the projection to state space is only partial, the initial conditions are completed by a trainable linear projection to the remained states, $\tilde{z}_0 = [\hat{z}_0; f_{init}(\hat{z}_0)]$. Otherwise, the initial projected state is used, $\tilde{z}_0 = \hat{z}_0$.
The system is then evolved according to the generative model (\ref{eq:gen-model}), using a random sample $\bs{\epsilon}$ from the posterior distribution. Using teacher forcing, every $\tau$-th step the simulated state is replaced by the estimated state,
\begin{align}
\tilde{z}_{t+1} &= 
\begin{cases}
\tanh( \ f(\tilde{z}_t) + B \epsilon_t \ )                      & \text{if } t \bmod \tau \neq 0, \\
\tanh( \ f([\hat{z}_t; T_{d_{\hat{z}}:d_z} \tilde{z}_t]) + B \epsilon_t \ ) & \text{if } t \bmod \tau = 0.
\end{cases}
\label{eq:teacher-forcing}
\end{align}
Here, $T_{d_{\hat{z}}:d_z}$ is a truncation matrix selecting the elements from the $d_{\hat{z}}$-th to the last $d_z$-th dimension; that is, the remaining states for which the state was not estimated are left to evolve freely.

For each sample $\bs{x}$, the loss function is composed from the reconstruction loss of $\bs{x}$ and $\bs{\hat{z}}$, and the KL term of the latent noise variables $\bs{\epsilon}$
\begin{equation}
  L = L^{\mathrm{rec}}_x + L^{\mathrm{rec}}_{\hat{z}} + L^{\mathrm{KL}} .
  \label{eq:loss}
\end{equation}

The reconstruction loss of the observation is
\begin{equation}
  L^{\mathrm{rec}}_x = \mathbb{E}_{\bs{\epsilon} \sim q(\bs{\epsilon} \mid \bs{x}, \hat{\bs{z}})} [-\log p(\bs{x} \mid \tilde{\bs{z}})],
  \label{eq:lrecx}
\end{equation}
where $p(\bs{x} \mid \tilde{\bs{z}}) = \prod_t p(x_t \mid g(\tilde{z}_t), \Sigma_\eta)$ using the observation function $g$ and covariance $\Sigma_{\eta}$ from (\ref{eq:gen-model}).
We treat the estimated states as additional observations with identity mapping, we find that such an approach helps to stabilize the training. Therefore, the reconstruction loss of the estimated partial states is similar to the observation reconstruction loss, apart from replacing the observation operator by the identity function on the estimated states:
\begin{equation}
  L^{\mathrm{rec}}_{\hat{z}} = \mathbb{E}_{\bs{\epsilon} \sim q(\bs{\epsilon} \mid \bs{x}, \hat{\bs{z}})} [- \log p(\hat{\bs{z}} \mid \tilde{\bs{z}})],
    \label{eq:lrecz}
\end{equation}
where $p(\hat{\bs{z}} \mid \tilde{\bs{z}}) = \prod_t p(\hat{z}_t \mid T_{1:d_{\hat{z}}} \, \tilde{z}_t, \Sigma_{\hat{z}})$ with $T_{1:d_{\hat{z}}}$ being the truncation matrix selecting the first $d_{\hat{z}}$ states, and the covariance matrix being diagonal and isotropic, $\Sigma_{\hat{z}} = \sigma_{\hat{z}}^2 I$.
Finally, the KL term in the loss follows the standard formulation of variational autoencoders, 
\begin{equation}
  L^{\mathrm{KL}} = D_{\mathrm{KL}} (q(\bs{\epsilon} \mid \bs{x}, \hat{\bs{z}}) \;\|\; p(\bs{\epsilon})) .
  \label{eq:kleps}
\end{equation}
We use a prior of standard normal distribution $p(\bs{\epsilon}) = N(\bs{\epsilon} \mid 0, I)$.
In addition, two regularization terms are added to the loss. First, to favour sparse observation models, it is L1 regularization on the weights of projection $g$.
Second, to encourage desired scale of the state trajectories, it is a regularization term for the scale and position of the estimated states.
The loss function is minimized using Adam optimizer.
Further details of the architecture and training are given in Sec.~\ref{sec:details-dpdsr}.

\subsection{Compared methods}

\textbf{Single projection DSR} (SPDSR) is a deterministic variant of DPDSR. It uses the same architecture and teacher forcing training method, but it assumes that there is no noise in the dynamical system. The noise encoder is therefore absent, and the KL divergence term does not appear in the loss function (\ref{eq:loss}).
\textbf{Generalized teacher forcing} \citep{HessEtAl23} is a method for reconstruction of deterministic dynamical system, and has been shown to outperform other deterministic methods.
We use an approach with time delay embedding (GTF-TD) where the observation is first projected to the state space so that the full state can be forced. We used PECUZAL method for the time delay embedding \citep{KraemerEtAl21}, and where this failed, we defaulted to a delay embedding with constant offset and predetermined dimension $d=8$.
\textbf{Deep Kalman Filter} (DKF) \citep{KrishnanEtAl15,KrishnanEtAl17} is a method for reconstruction of stochastic systems. Using an encoder, it estimates the states of the system, while the one-step stochastic prediction of the generative model forms the basis of the loss function. 
Our implementation here mirrors the architecture of our proposed method where possible (in the architecture of the encoder and of the generative model), with the main difference being the form of the loss function.
\textbf{Autoregressive LSTM} (AR-LSTM) \citep{Graves14} uses a standard LSTM network whose probabilistic output is fed back to the network at the next step, thus forming a stochastic dynamical system. During training, the original time series are used to feed the network, or can be replaced by the model generated output following the ideas of scheduled sampling \citep{BengioEtAl15}. 
\textbf{Recurrent State Space Model} \citep{HafnerEtAl19} combines a deterministic recurrent neural network with stochastic states as its generative model. During training, the stochastic states are estimated during rollout from the deterministic states and observations. We evaluate two variants: a single-step prediction variant (RSSM), and a multi-step variant with observation overshooting (RSSM-OO).
The generative models of all methods were approximately matched in number of parameters (Tab.~\ref{tab:params-gen}).

\section{Results}

\begin{figure}[t]
\begin{center}
  \includegraphics[width=1.0\textwidth]{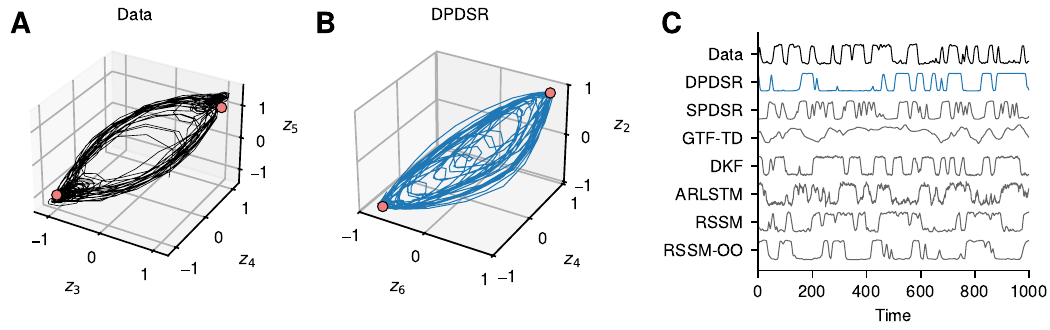}
  \caption{Results for the double well dataset.
    (A) Trajectory in the original state space of the model. The dots represent the stable fixed points of the model. 
    (B) Simulated trajectory in the state space of the DPDSR model (hand-picked dimensions). 
    (C) Original time series, and time series generated by the trained models. Here, and in all other figures, the time is represented in sample indices and not the original model time.
  }
\label{fig:doublewell}
\end{center}
\end{figure}

\paragraph{Example: Double well model}
First, we demonstrate the working of the methods on an example of stochastic dynamics: a noise driven double well model (Fig.~\ref{fig:doublewell}).
The double well dataset (Sec.~\ref{sec:doublewell}) is generated by numerical integration of a stochastic differential equation of bistable dynamics followed by four layers of exponential smoothing,
\begin{align*}
  \dot{z}_1 & = -z_1^3 + z_1 + \sigma \eta(t), \\
  \dot{z}_{i} & = \alpha (z_{i-1} - z_{i}) \quad  \text{ for } i \in \{2,3,4,5\}, \nonumber
\end{align*}
with parameters $\sigma$ scaling the noise amplitude and $\alpha$ being the temporal constant of exponential smoothing.
We assume only the last variable $z_5$ is observable.
The model has two stable fixed points, $z_i = \pm 1, \; i \in \{1,\ldots,5\}$.

As shown on Fig.~\ref{fig:doublewell}C, reconstruction methods based on deterministic dynamics do not perform well on this problem. A good time delay embedding cannot be found for time series generated by stochastic dynamics, and GTF-TD fails to reproduce the bistable nature of the time series.
The embedding-based method using deterministic dynamics (SPDSR) performs better and is able to reproduce the bimodal distribution of the data via chaotic dynamics, but a quantitative evaluation in the following section will reveal the inferior match compared to the stochastic method.
Among the stochastic methods, all of DPDSR, DKF and RSSM in both variants reproduce the bimodal nature of the data well.
Autoregressive LSTM performs better than the deterministic methods, but worse than embedding-based stochastic methods.

\paragraph{Quantitative evaluation}

For a more detailed evaluation of the DPDSR method, we test it on six datasets (Sec.~\ref{sec:datasets}):
1) Lorenz, the well-known classic example of deterministic chaos \citep{Lorenz63}. Even though training models of deterministic chaos is not the main target of the work, it is an important special case for stochastic methods too;
2) Cell cycle, a six-dimensional model of cell division cycle, generated by a deterministic chaotic model \citep{RomondEtAl99};
3) Double well, the dataset described above;
4) RNN, time series generated by a large recurrent neural network in chaotic regime \citep{SompolinskyEtAl88}, of which only one variable is observed. Due to the complexity of the model and minimal observation, we expect that the deterministic dynamics cannot be recovered, and stochastic model would prove to be a better alternative;
5) Neuron, an experimental dataset of somatic voltage of a rat pyramidal neuron driven by random and unknown stimulus current \citep{JolivetEtAl06};
6) ECG, an electrocardiogram signal of a healthy adult \citep{ReissEtAl19}.

\begin{figure}[t]
\begin{center}
 \includegraphics[width=\textwidth]{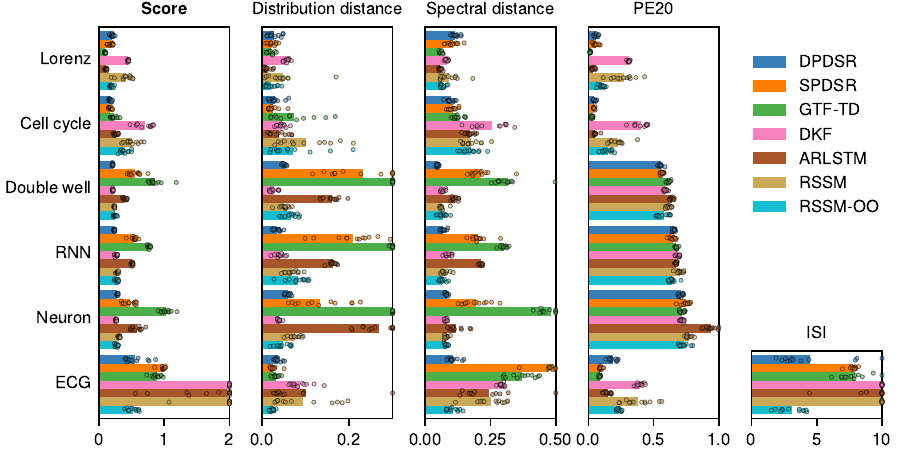}
  \caption{Reconstruction quality of the evaluated methods for the six datasets. The panels show the overall score, distribution distance $D_d$, spectral distance $D_s$, the 20-step prediction error $\text{PE}_{20}$, and the distance interspike interval distributions $D_\text{ISI}$ (lower = better for all measures).
    Each circle represent one of twelve initializations of the training method, and the bar shows the mean.
    For clearer visualization, the results were clipped to the shown interval.
  }
\label{fig:main-results-ext}
\end{center}
\end{figure}

We evaluate the quality of the reconstruction based on three criteria (Sec.~\ref{sec:eval-crit}). First, the distribution distance $D_d$: the similarity of the spatial distributions of the original and model-generated data. Specifically, we generate long time series with the trained model, collapse the data across time, and compare the distributions using Wasserstein distance.
Second, the spectral distance $D_s$: the similarity of the original and model-generated data in the frequency domain. We compute the power spectra of the original and model-generated data, and compare those using the Hellinger distance.
And third, the short-term prediction error $\text{PE}_{20}$ of the model-generated trajectories when starting from a state estimated from past data.
For the ECG dataset, we also include the distance of interspike intervals $D_\text{ISI}$ as an important measure of the reconstruction quality.
We summarize these measures in a cumulative score, which we define as a weighted sum of the described measures,
  ${S = w_1 D_d + w_2 D_s + w_3 \text{PE}_{20} + w_4 D_\text{ISI}}$.
Considering the different nature of the datasets, we set the weights differently for the different datasets in order to reflect the importance of the measures for each specific dataset (Sec.~\ref{sec:eval-crit}).
The results of the separate components are however also presented.

We train the models as described in Sec.~\ref{sec:methods} and Sec.~\ref{sec:methods-supp}.
For each method, we perform a parameter sweep over selected hyperparameters, and for each parameter combination  we train the model with sixteen random initializations, of which we discard four worst for each method.
We then select the best model as the one with the lowest average score across initializations.
The resulting scores for all problems are presented on Fig.~\ref{fig:main-results-ext}, and examples of generated time series are on Fig.~\ref{fig:time-series-all}.

We summarize the results in three main points.
First, for the datasets generated by low-dimensional deterministic models (Lorenz and Cell cycle), we see a good performance of deterministic models, either based on time delay embedding (GTF-TD) or trained projection (SPDSR). Our stochastic model is, however, competitive in all three measures.
DKF performs badly, mainly due to its limited prediction capacity.

Second, for the Double well, RNN, and Neuron datasets, where the time series are dominated by random (or seemingly random) effects on short time scales, the stochastic models DPDSR, DKF, and RSSM show the best performance.
The deterministic models (SPDSR and GTF) have to approximate the random transitions through deterministic chaos; the SPDSR projection method does it better than the GTF-TD method based on the time embedding.
We note that for these three dataset, the state of the art PECUZAL embedding failed and we used embedding with fixed delay and number of dimension instead. Thus, a worse performance of the time embedding method can be expected.
The stochastic AR-LSTM method stands in between the two groups.

Third, for the ECG dataset, several method are capable of learning a dynamical system that can generate the periodic ECG signal (Fig.~\ref{fig:time-series-all}).
However, the variations of the interspike intervals (ISI) in the data are not reproduced by the deterministic models SPDSR and GTF (Fig.~\ref{fig:main-results-ext}, rightmost column), leading also to higher spectral distance and higher overall score.
Of the stochastic methods, only our proposed DPDSR method and the multi-step RSSM-OO method can produce the variations in the interspike intervals. Of these two, the RSSM-OO method is more robust to weight initialization.

\paragraph{Internal dynamics of the trained models}

To better understand the behavior of the proposed DPDSR method, we now analyze the trained models using tools from dynamical systems theory.
Specifically, we look at the nature of the attractors of the deterministic part of the trained generative models.
For contrast, we compare the results with those from the deterministic SPDSR models. Apart from the absence of the noise, to SPDSR method is identical to the DPDSR in the architecture and training procedure. As such, it provides a useful comparison point for the differences of the stochastic and deterministic methods.

\begin{figure}[t]
\begin{center}
    \includegraphics[width=\textwidth]{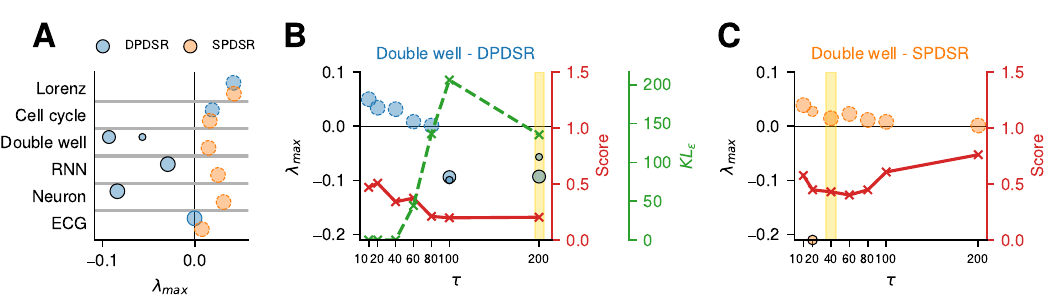}
  \caption{Analysis of the attractors of the trained models.
    (A) Maximal Lyapunov exponent $\lambda_{\text{max}}$ for the models trained on the six datasets. Each point correspond to one attractor. Dashed circle outline represents a chaotic attractor ($\lambda_{\text{max}} > 0$), colored solid outline represents a limit cycle, and black solid outline a fixed point. Size of the circle corresponds to the size of the basin of attraction.
    (B) Influence of the teacher forcing interval $\tau$ on the attractors in the DPDSR models. Circles show the maximal Lyapunov exponent as in (A). Solid red line shows the score (lower = better). Dashed green line shows the KL divergence of the estimated posterior distribution of the noise to the prior distribution $\mathrm{KL}_\epsilon = D_\mathrm{KL}(q(\bs{\epsilon} \mid \bs{x}) \;\|\; p(\bs{\epsilon}))$.
    The yellow band indicates the optimal $\tau$ value for the dataset.
    (C) As in (B), but for the deterministic SPDSR models.
  }
\label{fig:lyap}
\end{center}
\end{figure}

In Fig.~\ref{fig:lyap}A we show the maximal Lyapunov exponent of the attractors in the trained models for all six datasets.
For each dataset and method, we first detect the attractors by evolving the system from randomly initialized points, and then estimated the maximal Lyapunov exponent (Sec.~\ref{sec:attractors}).
The results show that the DPDSR method learns dynamical system exhibiting deterministic chaos for the Lorenz and Cell cycle dataset, both of which are indeed generated by a chaotic system.
As shown also on Fig.~\ref{fig:doublewell}, the model trained on the Double well problem has two stable fixed points (with noise driven transitions between them). The similar RNN problem results in a noise-driven fluctuations around a single fixed point. The variations in the interspike intervals in the ECG model are due to noisy fluctuations around a stable limit cycle.

In contrast, the best performing deterministic models trained with the SPDSR method exhibit chaotic dynamics in all cases. Indeed, in the absence of noise, deterministic chaos is the only option to model the random transitions of the Double well and RNN datasets, or spikes of the Neuron datasets.

Next, we explore the role of the teacher forcing interval $\tau$ on the performance and nature of the dynamics of the trained models.
The double well model (Fig.~\ref{fig:lyap}B,C) illustrates the phenomena that are to a large extent consistent across all datasets (Fig.~\ref{fig:lyap-all}).
In the stochastic DPDSR model, we observe that for smaller intervals ($\tau \leq 40$) the trained models contain chaotic attractors, and they do not rely on the noise. We quantify this by the KL divergence of the estimated posterior distribution of noise from the prior distribution, $\mathrm{KL}(q(\bs{\epsilon} \mid \bs{x}) \;\|\; p(\bs{\epsilon}))$, averaged across all samples. If the two distributions overlap, the KL divergence is zero, and no information is encoded in the posterior distribution. 
As the interval $\tau$ increases, the models transition into the regime of noise-driven dynamics with two stable fixed points. This is complemented by the increasing importance of the noise, so that the observed dynamics can be generated by noise-driven fluctuations.

Such behavior is mostly consistent for DPDSR models across datasets, with variations in the position of the transition into the noise-driven regime or the number and nature (fixed point or limit cycle) of the stable attractors.
However, the position of the optimal model can differ: it is located in the deterministic chaos regime for Lorenz and Cell cycle, or stochastic regime for Double well, RNN, Neuron, and ECG problems. 
In the deterministic SPDSR models, contrastingly, the stochastic regime cannot exist by construction. 
The models therefore mainly stay in the chaotic regime. The best performing models are found for lower teacher forcing intervals $\tau$, and the score worsens for increasing values.
This behavior matches the results reported by \citet{MikhaeilEtAl22} when training deterministic models using a teacher forcing scheme on the data from Lorenz system, forced Duffing oscillator, and empirical temperature time series.

\paragraph{Choice of the teacher forcing interval}
\label{sec:choosing-tau}

An important question for practical applications of the method is how to select the teacher forcing interval $\tau$.
In this work, we have performed a parameter sweep across a range of values. Such approach, while robust, is costly in terms of computational time.
Existing works proposed several approaches to similar issues in deterministic models.
\citet{MikhaeilEtAl22} used teacher forcing interval equal to the predictability time of a chaotic system
\begin{equation}
  \tau_\text{opt} = \frac{\log 2}{\lambda_\text{max}}
  \label{eq:tauopt}
\end{equation}
with $\lambda_\text{max}$ being the maximal Lyapunov exponent estimated ahead from the data; they showed that such estimates match closely the optimal values.
\citet{HessEtAl23} introduced an adaptive scheme where the parameter of generalized teacher forcing is updated during the training based on the product of Jacobians of the trained system, evaluated along the forced trajectory.

\begin{figure}[t]
\begin{center}
  \includegraphics[width=1\textwidth]{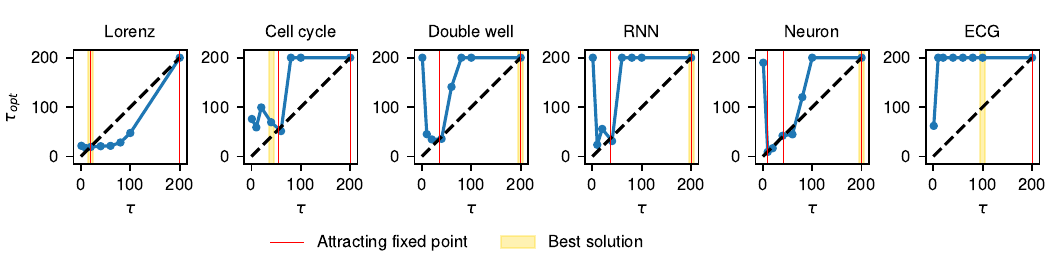}
  \caption{Optimal value of teacher forcing interval $\tau_\text{opt}$ across datasets.
    For each value of $\tau$, we calculated the maximal Lyapunov exponent $\lambda_\text{max}$ along the estimated trajectory using the final trained model.
    Following \citet{MikhaeilEtAl22}, we calculate $\tau_\text{opt} = \log 2 / \lambda_{\mathrm{max}}$ for $\lambda_\text{max} > 0$, and we set $\tau_\text{opt}$ equal to the maximal trajectory length (=200) otherwise.
    The red lines indicate the attracting fixed points of a relaxed fixed point iteration scheme $\tau_{i+1} = (1 - \alpha) \tau_i + \alpha \tau_{\text{opt},i}$ if ran on the visualized relation $\tau_\text{opt} (\tau)$ obtained from the final models, that is, they show the points of diagonal crossings with derivative smaller than 1. We assume linear interpolation between the evaluated points.
    The yellow line indicates the position of the best trained models, same as in Fig.~\ref{fig:lyap-all}.
    Note that here the Lyapunov exponents are evaluated along the estimated trajectories, and not for the deterministic attractors as in Fig.~\ref{fig:lyap-all}.
    For most datasets, two fixed points exist, one in the low $\tau$ (deterministic) regime, and one in the high $\tau$ (noise-driven) regime, indicating that the adaptive $\tau$ scheme might not robustly converge to the optimal solution.
  }
\label{fig:tau-opt}
\end{center}
\end{figure}

In the first approach, estimating the Lyapunov exponent from the data is performed under the assumption of deterministic chaotic system, and therefore is unsuitable for our purposes. However, a possible option would be to combine the two methods, that is, an adaptive scheme where at each step $i$ of the training we estimate the maximal Lyapunov exponent along the forced trajectory of our stochastic system, calculate the optimal interval $\tau_{\text{opt},i}$ using (\ref{eq:tauopt}), and adapt the parameter following a relaxed fixed point iteration scheme
\begin{equation}
  \label{eq:fpi}
  \tau_{i+1} = (1 - \alpha) \tau_i + \alpha \tau_{\text{opt},i}
\end{equation}
with relaxation parameter $\alpha \in (0, 1]$.

For a preliminary investigation if such approach is feasible, we reanalyze the results from our parameter sweep, and calculate $\tau_\text{opt}$ from the final trained models for each constant value of $\tau$ (Fig.~\ref{fig:tau-opt}).
Using this visualization we can see where the equation $\tau_\text{opt}(\tau) = \tau$ solved by scheme (\ref{eq:fpi}) has its attracting fixed points. These can provide an indication to which values of $\tau$ might the hypothetical adaptive scheme converge.
While it is not guaranteed that the adaptive scheme would converge to the same solutions as the scheme with fixed $\tau$, these results suggest that the scheme might not be sufficiently robust. On most datasets we see the existence of two attracting fixed points: one in the low range of $\tau$ in the deterministic regime with chaotic dynamics, and one in the noise-driven regime, where the negative Lyapunov exponent results in the maximal teacher forcing interval.
Although the best solution closely matches one of these fixed points for all but one problem, the adaptive scheme might not be able to correctly identify the optimal one.
We therefore conclude that for practical purposes where robust results are required the parameter sweep remains the best option, and we highlight of task of finding the optimal teacher forcing strategy in stochastic approaches as an interesting research problem for future studies.

\section{Conclusion and future directions}

In this work we have introduced a novel method for reconstructing stochastic dynamical models from the data.
The method is based on a double projection approach, where the time series of the observations are projected onto the estimates of the system states and the estimates of the driving noise using trained encoders.
The estimated system states are used for teacher forcing, while the driving noise is used as the latent variable in a variational autoencoder framework.
We benchmark the method on six test problems, demonstrating its performance for data generated by noise-driven models, deterministic chaotic models, and empirical data (Fig.~\ref{fig:main-results-ext}).
We then analyze the nature of the attractors in the learned dynamics in the examined problems, and evaluate the role of the teacher forcing interval $\tau$ on it (Fig.~\ref{fig:lyap} and Fig.~\ref{fig:lyap-all}). We identify the existence of two regimes occurring across the test problems: a deterministic regime for lower values of $\tau$ with chaotic attractors, and a noise-driven regime for higher values of $\tau$.

The proposed method has some notable shortcomings and limitations.
First, it is the dependency of the behavior of the trained model on the teacher forcing interval.
Performing parameter sweeps across a range of values to find the best performing model, while robust, is computationally demanding. 
As analyzed in Sec.~\ref{sec:choosing-tau}, our results indicate that an adaptive scheme for $\tau$, akin to the scheme of \citet{HessEtAl23}, might not be sufficient to converge to the optimally performing model.
The question of optimal strategy for setting the teacher forcing interval thus remains open for future studies.

The ability of a dynamical system reconstruction method to robustly learn dynamics with diverse timescales is of great importance for many problems, and improvements could be made to our proposed methods in this direction.
Gated variants of recurrent neural networks, most notably LSTM \citep{HochreiterSchmidhuber97} and GRU \citep{ChoEtAl14a}, were designed to deal with long-range temporal dependencies.
Other approaches with explicit time scale separation were suggested for DSR with deterministic models, often outperforming the traditional architectures.
Among the proposed approaches were: using multiple coupled RNNs operating with different temporal resolution \citep{LiuEtAl22,FarooqEtAl24},
regularization of the parameters of the neural network to introduce multiple time scales \citep{SchmidtEtAl21},
separating time scales using dynamic mode decomposition before applying the DSR algorithm \citep{BramburgerEtAl20},
or using echo state networks with different leak rates of leaky integrator neurons \citep{TanakaEtAl22}.
Applying these ideas to stochastic models could open a way to more robust methods for problems with disparate time scales.

\subsubsection*{Acknowledgments}

This project/research has received funding from the European Union’s Horizon Europe Programme under the Specific Grant Agreement No. 101137289 (Virtual Brain Twin Project), No. 101147319 (EBRAINS 2.0 Project), and No. 101057429 (project environMENTAL). This work has benefited from a government grant managed by the Agence Nationale de la Recherche (ANR) under the France 2030 program, reference ANR-22-PESN-0012.
Centre de Calcul Intensif d'Aix-Marseille is acknowledged for granting access to its high performance computing resources.


\newpage
\appendix

\renewcommand{\thefigure}{S\arabic{figure}}  
\setcounter{figure}{0}

\renewcommand{\thetable}{S\arabic{table}}  
\setcounter{table}{0} 

\section{Supplementary methods}

\subsection{Datasets}
\label{sec:datasets}

All datasets were generated or obtained and processed as described below, and all were normalized across time by subtracting the variable mean and dividing by the variable standard deviation.

\subsubsection{Lorenz system}

The Lorenz system \citep{Lorenz63} is a three-dimensional model exhibiting chaotic behavior, and is among the most used benchmarks in the dynamical system reconstruction field. It is described by three equations, 
\begin{align*}
  \dot{x} & = s (y - x),    \\
  \dot{y} & = rx - y - xz,  \\
  \dot{z} & = xy - bz,      \\ 
\end{align*}
with parameters $s = 10$, $r = 28$, $b = 2.667$. The system was simulated for $T = 10000$ using RK45 method implemented in Scipy library with the default relative and absolute tolerances of \num{e-3} and \num{e-6} respectively. The simulated time series are exported with sampling period $\Delta t = 0.05$, leaving \num{200000} time points, divided equally into the train and test time series. Only the first variable $x$ is considered to be observed.

\subsubsection{Cell cycle}

The model of cell division cycle \citep{RomondEtAl99} represents an example of deterministic chaos in a six-dimensional state space. It models the evolution of two coupled biochemical oscillators using six differential equations,
\begin{align*}
  \dot{C}_1 & =  v_{i1} \frac{K_{im1}}{K_{im1} + M_2} - v_{d1} X_1 \frac{C_1}{K_{d1} + C_1} - k_{d1} C_1 , \\
  \dot{M}_1 & =  V_1 \frac{1 - M_1}{K_1 + (1 - M_1)} - V_2 \frac{M_1}{K_2 + M_1}                         , \\
  \dot{X}_1 & =  V_3 \frac{1 - X_1}{K_3 + (1 - X_1)} - V_4 \frac{X_1}{K_4 + X_1}                         , \\
  \dot{C}_2 & =  v_{i2} \frac{K_{im2}}{K_{im2} + M_1} - v_{d2} X_2 \frac{C_2}{K_{d2} + C_2} - k_{d2} C_2 , \\
  \dot{M}_2 & =  U_1 \frac{1 - M_2}{H_1 + (1 - M_2)} - U_2 \frac{M_2}{H_2 + M_2}                         , \\
  \dot{X}_2 & =  U_3 \frac{1 - X_2}{H_3 + (1 - X_2)} - U_4 \frac{X_2}{H_4 + X_2}                         , \\ 
\end{align*}
with 
\begin{align*}
  V_1 & = \frac{C_1}{K_{c1} + C_1} V_{M1}, \quad V_3 = M_1 \cdot V_{M3}, \\
  U_1 & = \frac{C_2}{K_{c2} + C_2} U_{M1}, \quad U_3 = M_2 \cdot U_{M3}, \\   
\end{align*}
and $V_{M1} = U_{M1} = 0.3$, $v_{i1} = v_{i2} = 0.05$, $K_{1,2,3,4} = H_{1,2,3,4} = 0.01$, $V_2 = U_2 = 0.15$, $V_{M3} = U_{M3} = 0.1$, $V_4 = U_4 = 0.05$, $ K_{c1} = K_{c2} = 0.5$, $v_{d1} = v_{d2} = 0.025$, $K_{d1} = K_{d2} = 0.02$, $k_{d1} = k_{d2} = 0.001$. Following \citet{Gilpin21}, who included the model in a chaotic system collection, we set the value of the bifurcation parameter $K_{im1} = K_{im2} = 0.65$, for which a single chaotic attractor exist in the state space.

The system was simulated for $T = 800000$ using RK45 method implemented in Scipy library with the default relative and absolute tolerances of \num{e-3} and \num{e-6} respectively, and maximum time step \num{0.04}. The simulated time series are exported with sampling period $\Delta t = 5$, leaving \num{160000} time points, divided equally into the train and test time series. Only the first variable $C_1$ is considered to be observed.

\subsubsection{Double well}
\label{sec:doublewell}

The double well model represents a system with two fixed points and noise driven switches between the two basins of attractions. It is described by a cubic stochastic differential equation with fixed points at -1 and 1, followed by four exponential smoothing equations,
\begin{align*}
  \dot{z}_1 & = -z_1^3 + z_1 + \sigma \eta(t),   \\
  \dot{z}_i & = \alpha (z_{i-1} - z_i)   \quad  \text{ for } i \in \{2,3,4,5\},
\end{align*}
with $\alpha = 0.4$ and additive Gaussian noise $\eta(t)$ with variance $\sigma^2 = 0.2$. The last variable $z_5$ is considered the observation of the system. The system is simulated with Euler-Maruyama method with $\Delta t = 0.2$ for time $T = 400000$, and then downsampled by a factor of \num{10}, leaving \num{200000} time points with sampling period \num{2}, equally divided into a train and test set. 

\subsubsection{Chaotic RNN}

The chaotic recurrent neural network model \citep{SompolinskyEtAl88} describes the activity of a population of $n$ randomly connected neurons. 
The dynamics of the synaptic current of the $i$-th neuron is given by
\begin{align*}
  \dot{h}_i = -h_i + \sum_{j=1}^n J_{ij} \phi(h_j),
\end{align*}
where $\phi(h) = \tanh(h)$, and the connectivity matrix $\bs{J}$ contains independent random elements $J_{ij}$ following Gaussian distribution with mean 0 and variance $g / n^2$. We choose the factor $g = 2$, for which the model exhibits chaotic behavior.

We set the number of neurons $n = 1000$, and we solve the system using RK45 method implemented in Scipy library with the default relative and absolute tolerances of \num{e-3} and \num{e-6} respectively. We solve the system for $T = 100000$ and export with sampling period $\Delta t = 0.5$, leaving \num{200000} time points, divided equally into the train and test time series. Only the firing rate of the first neuron $\phi(h_1)$ is taken as observed. 

\subsubsection{Neuron}

The dataset represents the voltage time series of an \textit{in vitro} cortical pyramidal neuron from rat barrel cortex stimulated by a randomly generated fluctuating current \citep{RauchEtAl03,JolivetEtAl06}; the dataset was also used in a spike-timing prediction competition \citep{JolivetEtAl08} from which we obtained the data (\url{https://lcnwww.epfl.ch/gerstner/QuantNeuronMod2007/challenge.html}).
In the experiment, the neuron was stimulated by a current generated by an Ornstein-Uhlenbeck process. Although available in the data, here we have assumed that the input current is unknown, and aimed to estimate a stochastic dynamical model of both the neuron and the noise source.

From the available data, we have used only one recording (\texttt{00-0}). The signal was sampled at \qty{5}{\kHz} and contained \num{34000} data points, we have discarded the initial \num{600} and final \num{1200} time points where the stimulus current was not applied. The time series were smoothed with a Gaussian filter with $\sigma = \qty{0.2}{\ms}$ , normalized over time, and divided equally into the train and test time series of \num{16100} time points each.

\subsubsection{Electrocardiogram (ECG)}

The ECG signal captures the heart's electrical activity over time. We have used human ECG recording from the PPG-DaLiA dataset \citep{ReissEtAl19}, as preprocessed and used by \citet{HessEtAl23}. They performed smoothing and normalization of the time series, and used the signal of length \num{100000} time points at sampling rate \qty{700}{\Hz} (duration \qty{143}{\s}) for both training and test dataset. We have further downsampled the time series by a factor of 4 (\num{25000} time points, sampling rate \qty{175}{\Hz}, duration \qty{143}{\s} for each train and test time series) for reasons of reduced computational costs. 

For the sake of consistency with other datasets in this study, we have not used the time embedding provided by \citet{HessEtAl23}, but performed the time embedding ourselves (described below), although using the same embedding method as in the original study. Our approach resulted in 5-dimensional embedding, consistent with the results of \citet{HessEtAl23}.

\subsection{Time delay embedding}

For all time series we first attempted to perform the time delay embedding using the PECUZAL algorithm \citep{KraemerEtAl21} using the implementation in DelayEmbedding Julia module. Our settings allowed for possible delay values between 0 and 100 time points, the Theiler window $w$ was set to the first minimum of mutual information of the signal with itself, and we used the economy-mode for L-statistic computation, while all other arguments were set to default. The PECUZAL embedding was successfully achieved for the Lorenz system (dimension $d = 3$), cell cycle model ($d = 5$), and ECG signal ($d = 5$). The PECUZAL embedding failed for the double well system, chaotic RNN, and the neuron recording; in these cases we used a time delay embedding with repeated delays equal to the minimum of mutual information of the signal with itself, and hand-selected embedding dimension $d = 8$.

\subsection{Architecture and training details}
\label{sec:methods-supp}

For all methods, the same procedure was followed to train the models.
For each dataset, a parameter sweep was performed over selected hyperparameters (differing across methods), and for each parameter combination, four models were trained with four different random weight initializations.
The models were checkpointed during training. The models were evaluated on a separate test set in terms of the score, detailed below. The best parameter combination was chosen by the lowest score averaged across initializations.
Where only one model was used for a visualization, the best performing model from the four initializations of the best performing parameter combination was used unless stated otherwise.

\subsubsection{Double Projection Dynamical System Reconstruction (DPDSR)}
\label{sec:details-dpdsr}

The DPDSR method is described in the main text (Sec.~\ref{sec:methods}). Here we describe further details.
The regularization term for the observation function is an L1 regularization on the weights of the projection $g$ with coefficient $\alpha_g = 0.3$. The regularization term for the scale and position of the estimated states aims to weakly enforce the desired scale of the state trajectories. It has the form $\alpha_{\hat{z}} \, D_\mathrm{KL}( N(\mu_{\hat{z}}, \mathrm{diag}(\sigma_{\hat{z}}^2) \;\|\; N(0, I))$, where the mean and variances of the states are calculated across time and samples in a batch, but not features. Denoting the sample in a batch by a superscript, $\mu_{\hat{z}} = \mathbb{E}_{b,t} [\hat{z}_t^b]$ and $\sigma_{\hat{z}}^2 = \mathrm{Var}_{b,t} (\hat{z}^b_t)$. We are using value $\alpha_{\hat{z}} = 0.001$.

Tab.~\ref{tab:params-enc} shows the parameters of the encoder model, and
Tab.~\ref{tab:params-gen} shows the parameters of the generative model.
We divide the time series into chunks of $T = 300$, and use batch size 16.
To avoid the boundary effects of the convolutional encoders, we evaluate the losses (\ref{eq:lrecx}-\ref{eq:kleps}) on shortened trajectories $\bs{x}_{1+a:T-a}$ (and equally shortened $\bs{\epsilon}$, $\bs{\tilde{z}}$, and $\bs{\hat{z}}$) instead of the full trajectory $\bs{x} = \bs{x}_{1:T}$ with $a = 50$.

\begin{table*}[t]
\caption{Parameters of the encoder models of the DPDSR method.}
\label{tab:params-enc}
\vskip 0.15in
\begin{center}
\begin{small}
\begin{sc}
\begin{tabular}{llc}
\toprule
  State encoder & Number of layers & 7  \\
                & Kernel size      & 7  \\
                & Dilation         & [1,2,4,8,16,32,64] \\
                & Channels         & 24 \\
  Noise encoder & Number of layers & 7  \\
                & Kernel size      & 7  \\
                & Dilation         & [1,2,4,8,16,32,64] \\  
                & Channels         & 24 \\
                & LSTM state size  & 32 \\
\midrule
\end{tabular}
\end{sc}
\end{small}
\end{center}
\vskip -0.1in
\end{table*}

\begin{table*}[t]
\caption{Parameters of the generative models.}
\label{tab:params-gen}
\vskip 0.15in
\begin{center}
\begin{small}
\begin{sc}
\begin{tabular}{llccc}
\toprule
  &  & Lorenz & Cell cycle & Double well  \\
  &  &        &  ECG       & RNN          \\
  &  &        &            & Neuron       \\  
\midrule
  DPDSR & State size     & 5    & 8    & 8    \\
         & f hidden size    & 256  & 256  & 256  \\
         & g hidden size  & 32   & 32   & 32   \\  
         & \# parameters  & 3015 & 4650 & 4650 \\
  \midrule
  SPDSR & State size     & 5    & 8    & 8    \\
         & f hidden size    & 256  & 256  & 256  \\
         & g hidden size  & 32   & 32   & 32   \\  
         & \# parameters  & 3046 & 4681 & 4681 \\  
\midrule
  GTF-TD & State size    & 3    & 5    & 8    \\
         & f hidden size   & 424  & 384  & 256  \\
         & \# parameters & 2974 & 4234 & 4368 \\
\midrule
  DKF    & State size    & 5    & 8    &  8    \\
         & f hidden size   & 256  & 256  & 256   \\
         & \# parameters & 2832 & 4377 & 4377  \\
\midrule
  AR-LSTM & State size    & 24   & 32   & 32   \\
          & Initial conditions size & 5 & 8 &  8 \\
          & \# parameters & 2642 & 4546 & 4546 \\
\midrule
  RSSM    & Deterministic state size & 24 & 32 & 32 \\
          & Stochastic state size    &  4 &  4 &  4 \\
          & Hidden layers in 2-layer nets & 16 & 16 & 16 \\
          & Number of parameters & 3194 & 4938 & 4938 \\
\midrule 
\end{tabular}
\end{sc}
\end{small}
\end{center}
\vskip -0.1in
\end{table*}

For all test problems, the parameter grid exploration was performed over two parameters: teacher forcing interval $\tau \in [1, 10, 20, 40,60, 80, 100, 200]$, and observation noise variance $\log \sigma_\eta^2 \in [-4, -2, 0]$. The noise variance of the estimated states was set to $\log \sigma_{\hat{z}}^2 = \log \sigma_\eta^2 + 2$ to avoid introducing more free parameters and to reflect the secondary importance of reconstructing the estimated states compared to reconstructing the original observations. 

The loss function is minimized using Adam optimizer. The optimizer is ran for 30k batch evaluations, with learning rate starting at 0.001 and reduced by a factor of 0.3 at 10k and 20k steps. The reparameterization trick of variational autoencoders is used to sample from the posterior, with 4 samples used to evaluate the expectations in (\ref{eq:lrecx}) and (\ref{eq:lrecz}).
We perform gradient clipping with threshold 100.
The model is saved every 5000 iterations.

\paragraph{Causal encoder}
\label{sec:causal-encoder}

The DPDSR method uses encoders based on dilated convolutional neural networks.
These encoders are non-causal, that is, the estimated states $\hat{z}_t$ at time $t$ are computed from both past and future observations, and therefore $\hat{z}_t$ depends on all elements of $\bs{x}_{1:T}$.
Such approach allows to effectively gather information from the whole provided sample.
However, evaluating the predictive performance of such model is difficult, as the initial state would need to be estimated also from future data, invalidating the results.
For this reason, we have trained also an auxiliary causal encoder, which uses the same architecture of dilated convolutional networks, but with all connections to future observations set to zero, so that $\hat{z}_t$ depends only on $\bs{x}_{1:t}$.

The causal encoder is trained in parallel with the main model. In each iteration, the loss and the gradients of the main model are first computed and the update is performed. After that, an update of the causal encoder is performed. Denoting the non-causal encoder as $F$ and causal encoder as $F_c$, the loss function is given by
\begin{equation*}
 L^{ce} = \| F(\bs{x}) - F_c(\bs{x}) \|,
\end{equation*}
that is, the causal encoder is trained so its output matches the output of the non-causal encoder.
We use Adam optimizer with the same learning rate and same batch sizes as for the main model.
The causal encoder is used only for the prediction tasks, while for all other purposes we are using the non-causal variant.
The noise posterior distribution is always estimated using the non-causal encoder, since for the prediction tasks the noise is sampled from the prior and not the posterior distribution, and future information is thus not leaked through it.

\subsubsection{Single Projection Dynamical System Reconstruction (SPDSR)}

The architecture of DPDSR allows to consider a special case of a noise-free model by setting $B = 0$ in the generative model (\ref{eq:gen-model}). In this case, no noise is used in the simulations, and the noise encoder is not used; for this reason we call this special case a Single Projection Dynamical System Reconstruction (SPDSR).
In effect, SPDSR is a variant of sparse teacher forcing methods for training deterministic systems, with the teacher signal estimated using the encoder based on convolutional neural networks.
We use this variant as a useful comparison point, as it allows us to directly evaluate the effects of the stochastic formulation of DPDSR against a method equivalent in the architecture and training methods.

Indeed, the architecture and training of SPDSR is equal to DPDSR, with the difference that the noise encoder is not employed, and the KL divergence term of the loss function (\ref{eq:kleps}) is zero.
Same as for DPDSR, the parameter grid exploration was performed over teacher forcing interval $\tau \in [1, 10, 20, 40,60, 80, 100, 200]$, and observation noise variance $\log \sigma_\eta^2 \in [-4, -2, 0]$.
As for the DPDSR method, a secondary causal encoder (Sec.~\ref{sec:causal-encoder}) is trained and used for the prediction tasks.

\subsubsection{Generalized Teacher Forcing (GTF)}

Generalized teacher forcing is a training method developed for training deterministic models of chaotic dynamics, and has shown a superior performance on range of problems \citep{HessEtAl23}. We are using the method to compare our method with state of the art method for deterministic dynamical system reconstruction.
Strictly speaking, ``generalized teacher forcing'' refers only to the training method, but in this paper we use it as a shorthand for both the training method and the specific architecture of the generative model used in the original paper of \citet{HessEtAl23}.

The generative model has the form
\begin{equation*}
  z_t = A z_{t-1} + W_1 \sigma(W_2 z_{t-1} + h_2) + h_1
\end{equation*}
similar to our parameterization (\ref{eq:fz}) apart from the diagonal matrix $A$, the tanh transformation in our method, and absent noise.
The model is trained using a generalized teacher forcing scheme. In every step, the state is replaced by a linear interpolation between the simulated state $\tilde{z}_t$ and the data $z_t$,
\begin{equation}
  \label{eq:gtf}
  \hat{z}_t = (1 - \alpha) \tilde{z}_t + \alpha z_t,
\end{equation}
with coefficient $\alpha \in [0;1]$.
Such approach was shown theoretically to rectify to problem of exploding gradients in learning chaotic dynamics.

We tested a variant of this approach using time embedding for state space projection. 
In this approach, the full states of the system are estimated using time delay embedding (GTF-TD), and these states are used in the forcing scheme (\ref{eq:gtf}). The observations are then equal to the states, $x_t = z_t$. 
We used the original implementation in Julia language provided by authors (\url{https://github.com/DurstewitzLab/GTF-shPLRNN}).

We followed the example in \citep{HessEtAl23} to set the method parameters. 
We divided the time series into chunks of size $T = 200$ and used batch size 16.
The parameters were optimized using RADAM optimizer with 5000 epochs, 50 batches per epoch, and exponential decay schedule with initial and final learning rate $10^{-3}$ and $10^{-6}$ respectively.
Every 1000 epochs the model was saved and evaluated, and best performing model was kept.
For all test problems, the parameter sweep was performed across the value of teacher forcing parameter $\alpha \in [0.0, 0.1, 0.2, 0.3, 0.4, 0.5]$. 
The parameters of the generative model are given in Tab.~\ref{tab:params-gen}.

\subsubsection{Deep Kalman Filter (DKF)}

Deep Kalman Filter \citep{KrishnanEtAl15,KrishnanEtAl17} is based upon the framework of variational autoencoders applied to dynamical systems, where the states of the system $\bs{z}$ are considered to be the latent variables.
The generative model of DKF is
\begin{align*}
  z_{t} & = f(z_{t-1}) + \Sigma_\epsilon \epsilon_t , \\
  x_{t} & = g(z_{t})   + \Sigma_\eta \eta_t .
\end{align*}
We use the parameterization of the evolution function
\begin{equation*}
  f(z_t) = z_t + W_2 \sigma(W_1 z_t + b_1) + b_2 
\end{equation*}
and linear observation function $g$. This formulation differs from the one used in the DPDSR method (\ref{eq:fz}) by the absence of the tanh nonlinearity; we have found that this stabilization is not needed for training the DKF models.

During training, for an observation $\bs{x}$, DKF first projects the observation to a posterior distribution state trajectories $q(\bs{z} \mid \bs{x})$.
Then a state trajectory is sampled from the posterior, $\bs{z} \sim q(\bs{z} \mid \bs{x})$, and the loss function for the data point is calculated as
\begin{equation*}
  L(\bs{x}) = D_\mathrm{KL} ( q(\bs{z} \mid \bs{x}) \;\|\; p(\bs{z}) ) - \mathbb{E}_{\bs{z} \sim q(\bs{z} | \bs{x})} \left[ \log p(\bs{x} \mid \bs{z}) \right].
\end{equation*}
The prior for the system states is
\[
  p(\bs{z}) = p(z_0) \prod_{t=1}^T p(z_{t+1} \mid z_t),
\]
with $p(z_0) = N(z_0 \mid 0, I)$ and $p(z_{t+1} \mid z_t) = N(z_{t+1} \mid f(z_t), \Sigma_\epsilon)$.
The probability in the reconstruction loss is
\[
  p(\bs{x} \mid \bs{z}) = \prod_{t=1}^T N(x_t \mid g(z_t), \Sigma_\eta).
\]
The system and observation noise covariance matrices were assumed to be diagonal and isotropic, $\Sigma_\epsilon = \sigma_\epsilon^2 I$ and $\Sigma_\eta = \sigma_\eta^2 I$.
Compared to our method, the evolution of the dynamical system is represented in the prior term in the KL divergence and the observation function in the reconstruction loss, while in our method both are represented in the reconstruction loss.
The disadvantage of the method is that it does not allow to evolve the system more than one step from the estimated states, leading to reduced capacity to learn long-term dependencies.

In the experiments, we are using the same architecture for the state encoder as for the noise encoder in the DPDSR method, that is, the encoder is composed of a stack of dilated convolutional networks followed by an autoregressive LSTM network. The same parameters as for the DPDSR noise encoder are used (Tab.~\ref{tab:params-enc}).
The loss function is minimized using Adam optimizer. The optimizer is ran for 30k batch evaluations, with learning rate starting at 0.001 and reduced by a factor of 0.3 at 10k and 20k steps. The models were saved every 5k steps.
The parameter sweep was performed over the observation noise variance, $\log \sigma_\eta^2 \in [-4, -2, 0]$ and initial values of the system noise variance, $\log \sigma_\epsilon^2 \in [-8, -6, -4, -2]$.  
As for the DPDSR method, a secondary causal encoder (Sec.~\ref{sec:causal-encoder}) is trained and used for the prediction tasks.

\subsubsection{Autoregressive LSTM (AR-LSTM)}

Long short-term memory (LSTM) network \citep{HochreiterSchmidhuber97} is an established architecture of recurrent neural networks designed to handle long-term dependencies in the input data. Autoregressive LSTM model \citep{Graves14} represents one approach to introduce stochasticity in the standard LSTM network.
The generative model is given by
\begin{align}
  h_t, c_t & = \mathrm{LSTM}(h_{t-1}, c_{t-1}, x_{t-1}), \label{eq:gen-arlstm} \\
       x_t & \sim N(\mu_t, \sigma_t^2) \notag,     
\end{align}
where $h_t$ and $c_t$ are the hidden state and cell state vectors at time $t$, and $\mathrm{LSTM}(h, c, x)$ represents the standard LSTM cell.
The observation $x_t$ at each step are drawn from a normal distribution, and fed back to LSTM as an input in the next step.
The parameters of the normal distribution are computed by a linear projection from the hidden state
\begin{equation*}
  [\mu_t; \log \sigma_t^2] = A h_t + b.
\end{equation*}

For training the model, the timeseries are split into chunks $\bs{x}$ with length $T = T_{\mathrm{past}} + T_{\mathrm{pred}}$.
The initial conditions $[h_0; c_0; x_0]$ are first estimated from the past observation of length $x_{1:T_{\mathrm{past}}}$.
This is done using a stack of dilated convolutional neural networks, mirroring the architecture of the state encoder in our method.
From the last layer and the last element in time the low-dimensional representation of the initial conditions $z_0 \in \R^d$ is computed via linear projection.
Then the full dimension initial conditions are computed through two layer MLP with ReLU nonlinearity.

From the initial conditions the system is evolved according to (\ref{eq:gen-arlstm}) to obtain the simulated observations $\tilde{\bs{x}}_{T_{\mathrm{past}+1}:T}$ and means and variances $\bs{\mu}_{T_{\mathrm{past}+1}:T}, \bs{\sigma}_{T_{\mathrm{past}+1}:T}$. Using the principle of scheduled sampling, during the evolution the observations entering the LSTM cell are either randomly sampled from the last step prediction ($\tilde{x}_t \sim N(\mu_t, \sigma_t^2)$) with probability $\gamma$, or replaced by the data $x_t$ with probability $1 - \gamma$.

The loss function for one sample $\bs{x}$ is given by
\begin{equation*}
  L(\bs{x}) = - \log p( \bs{x}_{T_{\mathrm{past}+1}:T} \mid \bs{\mu}_{T_{\mathrm{past}+1}:T}, \bs{\sigma}_{T_{\mathrm{past}+1}:T}^2) .
\end{equation*}
The loss function is minimized using Adam optimizer. The optimizer is ran for 30k batch evaluations, with learning rate starting at 0.001 and reduced by a factor of 0.3 at 10k and 20k steps. The models were saved every 5k steps.
The parameter sweep was performed for parameter $\gamma \in [0., 0.2, 0.4, 0.6, 0.8, 1.0]$ and for prediction length $T_\mathrm{pred} \in [20,50,100,200]$.

\subsection{Recurrent State Space Model (RSSM)}

The recurrent state space model \citep{HafnerEtAl19} combines a deterministic recurrent neural network with stochastic states. The generative model is given by the following relations:
\begin{alignat*}{2}
  & \textrm{Deterministic state model: } && z_t  = f(z_{t-1}, s_{t-1}),     \\
  & \textrm{Stochastic state model: }    && s_t  \sim p(s_t \mid z_t),      \\
  & \textrm{Observation model: }         && x_t  \sim p(x_t \mid s_t, z_t),
\end{alignat*}
with $z_t$ being the deterministic states, $s_t$ the stochastic states, and $x_t$ the observations.
The deterministic model is implemented using a GRU network \citep{ChoEtAl14a}, the stochastic transition model with a two-layer fully connected network projecting to the parameters of a diagonal Gaussian distribution, and the same architecture is used for the observation model.
The stochastic states are estimated using the encoder $q(\bs{s}_{1:T} \mid \bs{x}_{1:T}) = \prod_{t=1}^T q(s_t \mid h_t, r_t)$, where $r_t$ are features extracted from the observations, $r_t = g(\bs{x}_{1:T})$, using the same dilated convolutional network architecture as for the DPDSR model. The mapping from the features and deterministic states is again implemented using a two-layer fully connected network. Finally, the initial deterministic state $h_0 = h_0(r_0)$ is computed from the features, again using a two-layer fully connected network.
The parameters of the networks are optimized by minimizing the model ELBO,
\begin{equation}
  \label{eq:loss-rssm}
  L(\bs{x}) = \mathbb{E}_{q(\bs{s} \mid \bs{x})} \left[ \sum_{t=1}^T \log p(x_t \mid s_t, z_t) \right] - \mathbb{E}_{q(\bs{s} \mid \bs{x})} \left[ \sum_{t=1}^T \mathrm{KL} \left( q(s_t \mid h_t, r_t)  \;\|\; p(s_t \mid s_{t-1}) \right) \right].
\end{equation}

For the observation overshooting, or multi-step prediction, we use the equivalent adaptation of the cost function as in \citep{HafnerEtAl19}, except that the $d$-step prediction distribution $p(s_t | s_{t-d})$ is used in the reconstruction term (first term in (\ref{eq:loss-rssm})) and not the second term as in latent overshooting of \citep{HafnerEtAl19}. The generative model and the encoder networks otherwise remain the same.
For the overshooting approach we have performed parameter sweep over the maximal overshooting distance $d \in [3, 10, 20, 40 100]$ and we selected the best performing model
We have not evaluated latent overshooting approach, as its main motivation in \citep{HafnerEtAl19} was computational cost of the observation model which is comparatively small in our conditions. 

For all variants, the loss function is minimized using Adam optimizer. The optimizer is ran for 30k batch evaluations, with learning rate starting at 0.001 and reduced by a factor of 0.3 at 10k and 20k steps. The models were saved every 5k steps.

\subsection{Evaluation criteria}
\label{sec:eval-crit}

Dynamical system reconstruction aims at training models that can robustly reproduce the temporal patterns observed in the training data on long-term scale. In this spirit we evaluate the models using two measures evaluating long-term behavior, distribution distance $D_d$ and spectral distance $D_s$, and one measure of short-term prediction capacity, 20-step prediction error $\text{PE}_{20}$.
The first two are evaluated by a comparison of long model-generated time series with the original data.
To generate the data, we take a point on the embedded state trajectory (via trained projection embedding method, or time delay embedding depending on the model). Using this point as initial conditions, we evolve the system for 40000 steps, using random noise for stochastic models. 

The distribution distance $D_d$ measures the similarity of the distribution of the original and generated data in the observation space. To calculate it, we take the original and simulated data, collapse them across time, and compare the distributions using the Wasserstein distance (also known as Earth mover's distance).
Loosely speaking, the Wasserstein distance correspond to the cost of reshaping one distribution into the other by transporting the mass. In one dimension, the Wasserstein distance between two probability distributions $u$ and $v$ with cumulative distribution functions $F_u$ and $F_v$ is defined as:
\begin{equation}
  D_d(u, v) = \int_{-\infty}^{\infty} \left| F_u(x) - F_v(x) \right| \, dx .
  \label{eq:wasserstein}
\end{equation}
We use the SciPy implementation for computations.

The spectral distance $D_s$ measures the similarity of the long time series in frequency space.
To calculate it, we compute the power spectral density of the original and simulated data using Welch's method (using the SciPy implementation) with segment length equal to 4096 points.
We smooth the frequency spectra using a Gaussian filter with $\sigma = 2$ time points, normalize them, and calculate their Hellinger distance. The Hellinger distance for two discrete distributions $U$ and $V$ is given by
\begin{equation*}
  D_s(u,v) = \frac{1}{\sqrt{2}} \sqrt{\sum_i \left( \sqrt{u_i} - \sqrt{v_i} \right) }.
\end{equation*}

We also consider a measure of short term prediction capability, the $n$-step prediction error.
For data chunk $\bs{x} = (x_1, x_2, \ldots, x_T)$ we use the first $k$ time points to estimate the latent state at time $k$-th step. We then repeatedly simulate the next $n$ steps to with random noise obtain predictions $\tilde{\bs{x}} = (\tilde{x}_{k+1}, \ldots, \tilde{x}_{k+n})$.
The prediction error is then
\begin{equation*}
  \text{PE}_{n} = \frac{1}{n} \sum_{i=1}^n \| x_{k+i} - \tilde{x}_{k+i} \|,
\end{equation*}
which we average over 20 random noise samples (for probabilistic models only) and 2000 chunks from the test dataset.

For the ECG dataset, we also measure the distance between the original and simulated distributions of the interspike intervals $D_{\text{ISI}}$.
To do so, use the long time series generated as described above, and we detect the peaks in the signal using the SciPy \texttt{find\_peaks} tool with height~2 and prominence~1.
We then then set $D_{\text{ISI}}$ to be the Wasserstein distance (\ref{eq:wasserstein}) between the ISI distributions from the data and simulated signal.

We calculate the overall score as weighted sum of the distribution distance $D_d$, spectral distance $D_s$, 20-step prediction error $\text{PE}_{20}$, and (for ECG) the distance of ISI distributions $D_\text{ISI}$.
\begin{equation*}
  S = w_1 D_d + w_2 D_s + w_3 \text{PE}_{20} + w_4 D_\text{ISI}.
\end{equation*}
Given the different nature of the datasets, we set the weights differently across datasets.
For the Lorenz and Cell datasets, we use $\bs{w} = (w_1, w_2, w_3, w_4) = (1,1,1,0)$.
For the Double well, RNN, and Neuron datasets, where the predictability is lower, we reduce the weight on $\text{PE}_{20}$, $\bs{w} = (1,1,1,0.2,0)$.
For the ECG dataset, we include the distance of interspike intervals, $D_\text{ISI}$ as an important measure of the reconstruction quality, $\bs{w} = (1,1,1,0.05)$. The weight is chosen lower due to large magnitude of unnormalized values.
All measures are evaluated on a test dataset which was not used for training the model.

\subsection{Analysis of the attractors}
\label{sec:attractors}

To identify the attractors in the state space of the model, we randomly select 100 points on the state space trajectory obtained by projecting the training data into the state space using the trained encoder.
We then simulate the system forward for $T_{\text{warmup}} + T$ steps with $T_{\text{warmup}} = 1000$ and $T = 20000$.
For stochastic models, we set the noise in the simulation to zero.
After discarding the first $T_{\text{warmup}}$ steps to allow the transients to decay, the remaining trajectories are analyzed to detect distinct attractors. For each trajectory we compare it against previously identified attractors by computing pairwise distances between trajectory points.
Specifically, for each candidate new attractor trajectory ${\bs{z}^a \in \R^{T \times d_z}}$ and an already identified attractor ${\bs{z}^b \in \R^{T \times d_z}}$, we compute two way distances between the attractors,
\begin{align*}
  d_{a \rightarrow b} & = \mathrm{Q}_{t_a} \left( \min_{t_b} \| z^a_{t_a} - z^b_{t_b} \|, 0.8 \right), \\
  d_{b \rightarrow a} & = \mathrm{Q}_{t_b} \left( \min_{t_a} \| z^a_{t_a} - z^b_{t_b} \|, 0.8 \right),
\end{align*}
where $\mathrm{Q} \left( \cdot, q \right)$ denotes the $q$-th quantile. 
We define the attractor distance as ${d_{a,b} = \max(d_{a \rightarrow b}, d_{b \rightarrow a})}$, and consider the attractors to be distinct if $d_{a,b} > \text{tol}$, with $\text{tol} = 10^{-5}$ if the already identified attractor $b$ is a fixed point, and $\text{tol} = 10^{-1}$ for a limit cycle or a chaotic attractor.
We use the quantile instead of maximum for computing the distances, and relatively high tolerances, both for more robust solution when dealing with trajectories with finite length; this is at a cost of possibly conflating close attractors.

For each new attractor we calculate the maximal Lyapunov exponent $\lambda_{\max}$ numerically \citep{Sprott03}.
In the algorithm, we take a point on the attractor and a point with a small perturbation. We repeatedly advance the trajectories from the initial points using the deterministic dynamics, while rescaling the deviation of the perturbed trajectory to its original norm at every step.
We advance the system for 1000 time steps for, and calculate  $\lambda_{\max}$ as an average from the estimates from all steps.
We consider the attractor chaotic if $\lambda_{\max} > 0$, limit cycle if $\lambda_{\max} \leq 0$ and the trajectory does not converge to a single point (with $\text{tol} = 10^{-5}$), and fixed point otherwise.

\section{Supplementary results}

\subsection{Architecture variations}

\subsubsection{Number of parameters in the generative model}

We investigated how does the performance of the stochastic DPDSR method change when the number of parameters in the generative model is increased or decreased. In particular, we aimed at a comparison with the behavior of its deterministic modification: the SPDSR method.
We considered three problems, the Cell, Double well, and ECG datasets.
For each, we trained stochastic and deterministic models with 16 to 2048 units in the hidden layer of the generative model (\ref{eq:fz}), and evaluated the models as before. 

\begin{figure}[t]
\begin{center}
  \includegraphics[width=0.9\textwidth]{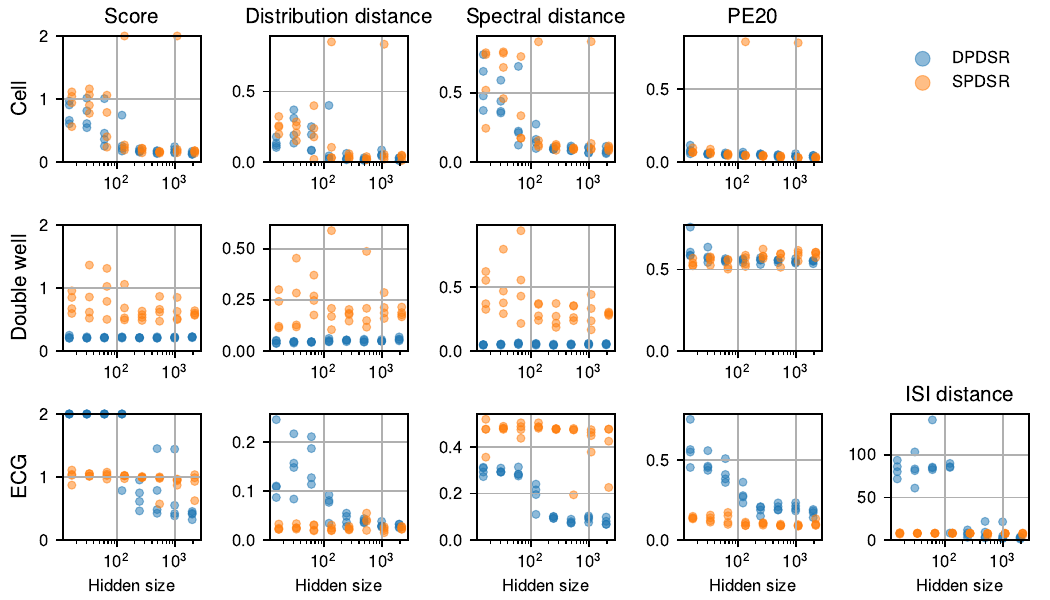}
  \caption{Effect of the number of units in the hidden layers.
    The effect on the model quality was investigated for three datasets (rows), the score and the separate measures of reconstruction quality are in the columns.
    For clearer visualization, the results were clipped at the upper limit of the shown range.
  }
\label{fig:nparams}
\end{center}
\end{figure}

Fig.~\ref{fig:nparams} shows that the behavior of the method depends on the dataset. For the Cell dataset, generated by a low-dimensional deterministic model, both DPDSR and SPDSR behave similarly, with increasing performance for increasing number of parameters.
For the Double well dataset, generated by a noise-driven model, the stochastic model performs well even for minimal number of parameters, and consistently outperforms the deterministic model.
For the ECG dataset we see that the deterministic models perform equally well across the range of parameters. The stochastic models, however, perform considerably worse for models with less than 100 units in the hidden layer, but outperform the deterministic models for larger hidden layers.

\subsubsection{Encoder architecture}

\begin{figure}[t]
\begin{center}
  \includegraphics[width=0.9\textwidth]{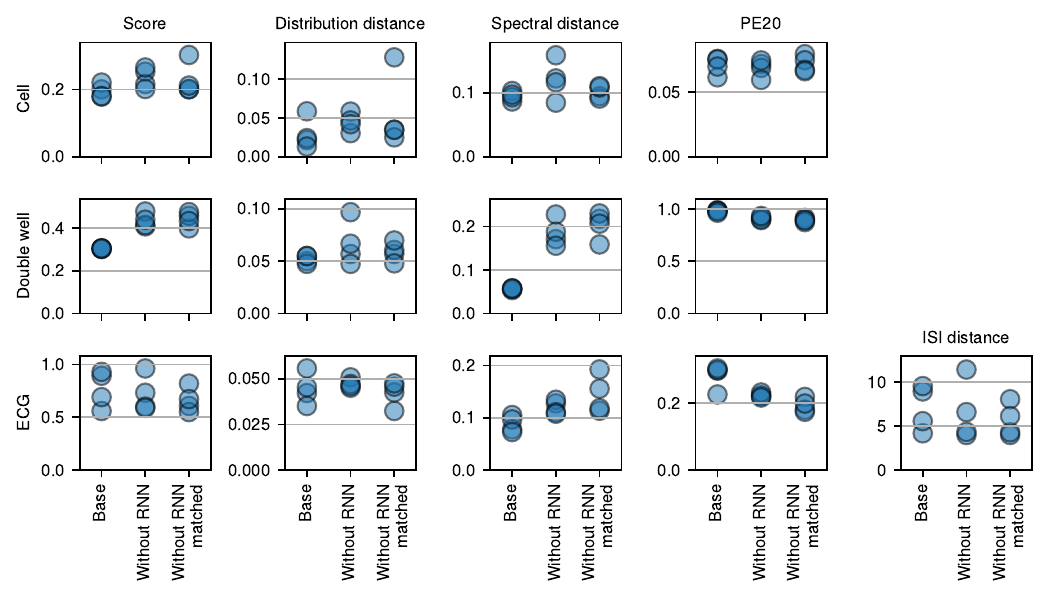}
  \caption{Role of the encoder architecture. For three datasets, three variations of the DPDSR model architecture were investigated: the baseline, a variation with the autoregressive RNN in the noise encoder removed, and a variation with the autoregressive RNN removed, but increased number of parameters in the convolutional neural network to match the total parameters.
    For each variant, four models with different random weight initializations were trained.
  }
\label{fig:encoder}
\end{center}
\end{figure}

We further investigated the role of the encoder architecture on the performance of the DPDSR method (Fig.~\ref{fig:encoder}). 
Specifically, we compared the baseline architecture with two modifications: First, a variation with a noise encoder with the autoregressive RNN removed, and with the parameters otherwise kept equal as described in Tab.~\ref{tab:params-enc}.
Second, a variation with a autogressive RNN in noise encoder also removed, but with increased number of channels (26 instead of 24) in the dilated convolutional neural network to keep the total number of parameters approximately equal.
Degradation in performance with the removed RNN can be seen for the Double well problem, while for the Cell and ECG problems the modification have little impact.

\begin{figure}[ht]
\begin{center}
    \includegraphics[width=\textwidth]{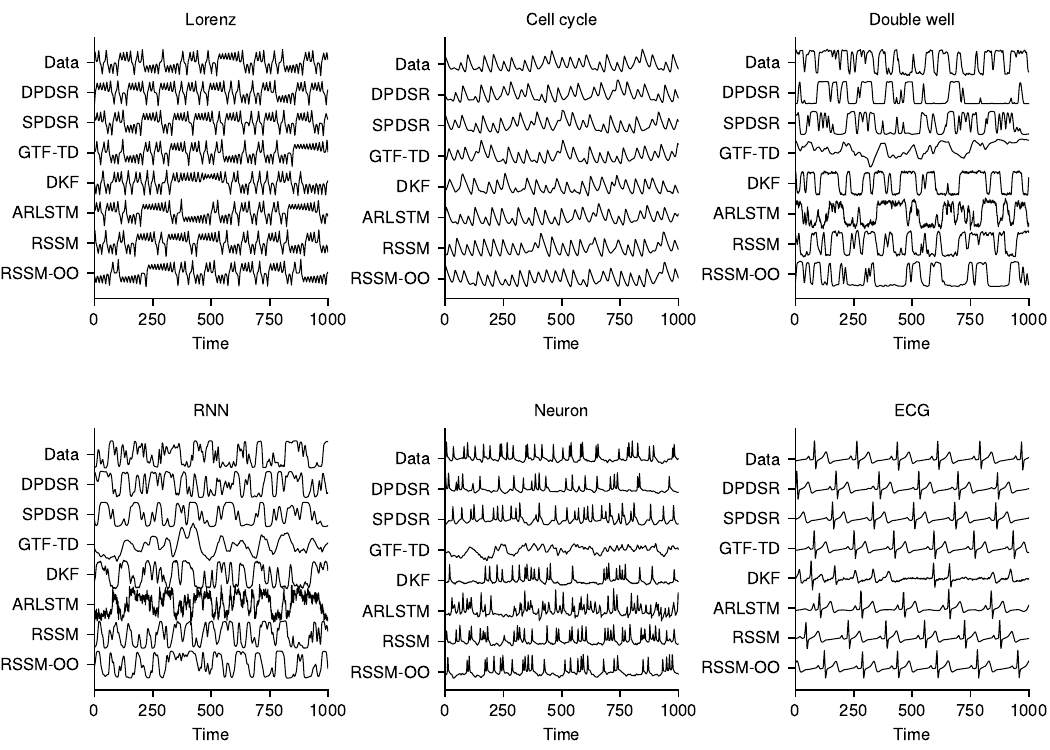}
  \caption{Examples of generated time series for all test problems and evaluated methods.
    The time is represented in sample indices and not the original model or real time.}
\label{fig:time-series-all}
\end{center}
\end{figure}

\begin{figure}[ht]
\begin{center}
  \includegraphics[width=1\textwidth]{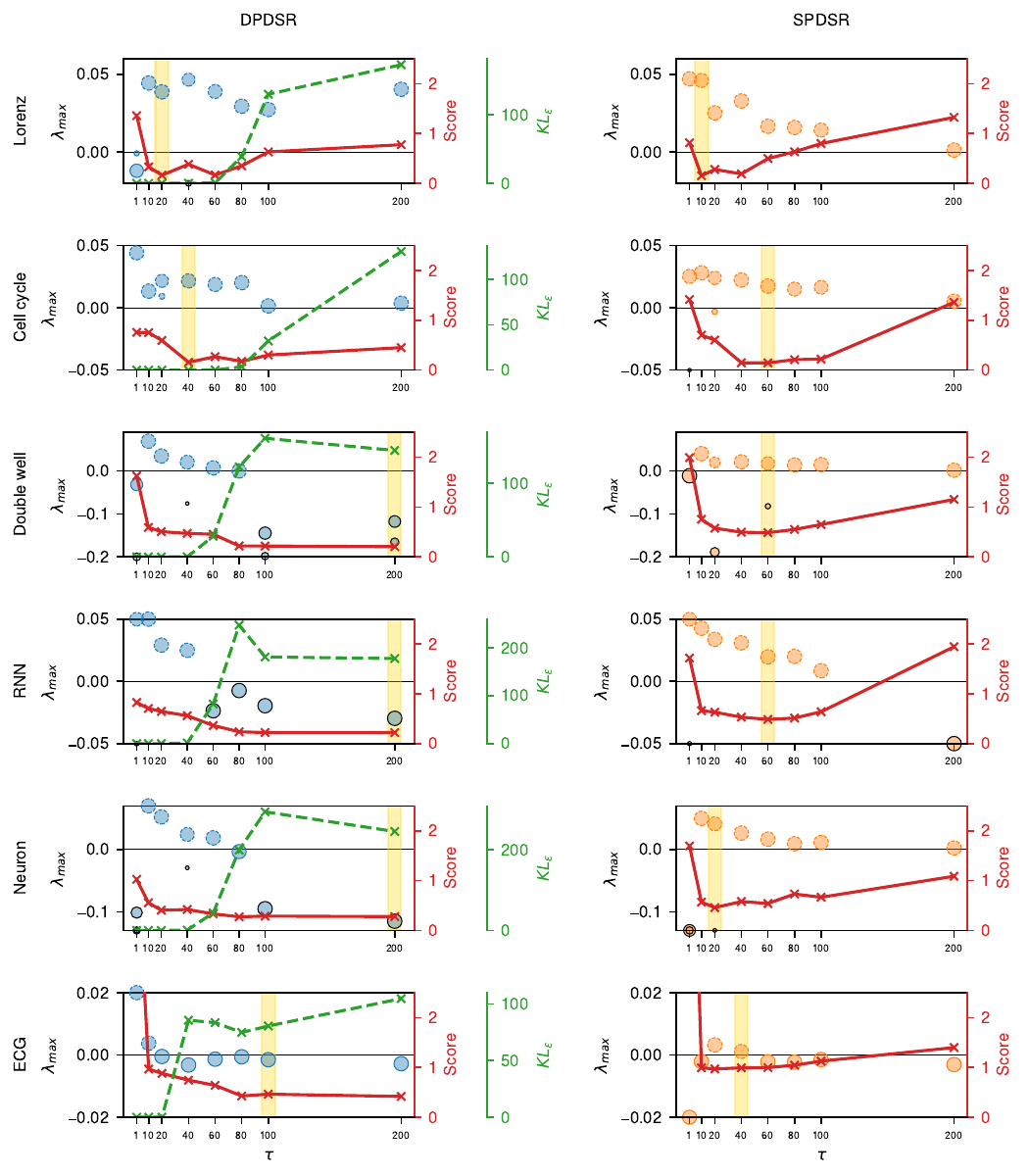}
  \caption{Analysis of the attractors of the trained models for all datasets, extending Fig.~\ref{fig:lyap}.
    Left column: DPDSR models (stochastic); right column: SPDSR models (deterministic).
    Circles show the maximal Lyapunov exponent.
    Dashed circle outline represents a chaotic attractor ($\lambda_{\text{max}} > 0$), colored solid outline represents a limit cycle, and black solid outline a fixed point. Size of the circle corresponds to the size of the basin of attraction.
    Solid red line shows the score (lower = better).
    Dashed green line shows the KL divergence of the estimated posterior distribution of the noise to the prior distribution $\mathrm{KL}_\epsilon = D_\mathrm{KL}(q(\bs{\epsilon} \mid \bs{x}) \;\|\; p(\bs{\epsilon}))$.
    The yellow band indicates the optimal $\tau$ value for the dataset; note that the optimal $\tau$ value is selected based on the mean score across all initializations, while for other visualizations we plot only the best model.
    For clearer visualization, the positions of the Lyapunov exponents are clipped to the shown interval.
  }
\label{fig:lyap-all}
\end{center}
\end{figure}

\end{document}